\documentclass[preprint,1p,12pt]{elsarticle}

\usepackage{amssymb}
\usepackage{amsmath}
\usepackage{subfigure}

\usepackage{times}    % Integrate Times for here
\usepackage{xspace}
\usepackage[dvipsnames]{xcolor}
\usepackage{graphicx}
\usepackage{amssymb}
\usepackage{amsmath}
\usepackage{mathrsfs}
\usepackage{pifont}
\usepackage{booktabs}
\usepackage[format=plain,labelformat=simple,labelsep=period,font=small,compatibility=false]{caption}
\usepackage[font=footnotesize,skip=3pt,subrefformat=parens]{subcaption}
\usepackage{etoolbox}
\usepackage{svg}
\usepackage[normalem]{ulem}
\usepackage{xurl}

\begin{document}

\begin{frontmatter}


% \author{Name\corref{cor1}\fnref{label2}}
% \affiliation{organization={},

\author[uniba]{Pasquale De Marinis\corref{cor1}}
\ead{pasquale.demarinis@uniba.it}

\author[jads]{Pieter M. Blok}
\ead{p.m.blok@tue.nl}

\author[jads,tu]{Uzay Kaymak}
\ead{u.kaymak@ieee.org}

\author[jads]{Rogier Brussee}
\ead{r.brussee@jads.nl}

\author[uniba]{Gennaro Vessio}
\ead{gennaro.vessio@uniba.it}

\author[uniba]{Giovanna Castellano}
\ead{giovanna.castellano@uniba.it}

\cortext[cor1]{Corresponding author}

\affiliation[uniba]{
  organization={Department of Computer Science, University of Bari Aldo Moro},
  addressline={Via Orabona, 4},
  city={Bari},
  postcode={70125},
  country={Italy}
}

\affiliation[jads]{
  organization={Jheronimus Academy of Data Science (JADS)},
  addressline={Sint Janssingel 92},
  city={'s-Hertogenbosch},
  postcode={5211 DA},
  country={The Netherlands}
}

\affiliation[tu]{
  organization={Eindhoven University of Technology (TU/e)},
  addressline={Het Eeuwsel 53},
  city={Eindhoven},
  postcode={5612 AZ},
  country={The Netherlands}
}

\title{DistillFSS: Synthesizing Few-Shot Knowledge into a Lightweight Segmentation Model}

% \affiliation[label1]{organization={},

% \affiliation[label2]{organization={},

% \author{} %% Author name

% \affiliation{organization={},%Department and Organization

%% Abstract
\begin{abstract}
Cross-Domain Few-Shot Semantic Segmentation (CD-FSS) seeks to segment unknown classes in unseen domains using only a few annotated examples. This setting is inherently challenging: source and target domains exhibit substantial distribution shifts, label spaces are disjoint, and support images are scarce--making standard episodic methods unreliable and computationally demanding at test time. To address these constraints, we propose DistillFSS, a framework that embeds support-set knowledge directly into a model's parameters through a teacher--student distillation process. By \textit{internalizing} few-shot reasoning into a dedicated layer within the student network, DistillFSS eliminates the need for support images at test time, enabling fast, lightweight inference, while allowing efficient extension to novel classes in unseen domains through rapid teacher-driven specialization. Combined with fine-tuning, the approach scales efficiently to large support sets and significantly reduces computational overhead. To evaluate the framework under realistic conditions, we introduce a new CD-FSS benchmark spanning medical imaging, industrial inspection, and remote sensing, with disjoint label spaces and variable support sizes. Experiments show that DistillFSS matches or surpasses state-of-the-art baselines, particularly in multi-class and multi-shot scenarios, while offering substantial efficiency gains. The code is available at \url{https://github.com/pasqualedem/DistillFSS}.
\end{abstract}


%% Keywords

\end{frontmatter}



\section{Introduction}
\label{sec:intro}

Deep neural networks for semantic segmentation, such as FCN \cite{longFullyConvolutionalNetworks2015}, DeepLab \cite{chenDeepLabSemanticImage2018}, and PSPNet \cite{zhaoPyramidSceneParsing2017}, have achieved remarkable success across a wide range of visual tasks. However, these models typically rely on large amounts of pixel-level annotations, which are cumbersome, costly, and time-consuming to obtain. This annotation burden becomes even more pronounced in specialized application domains---including medical imaging, industrial inspection, and satellite imagery---where data acquisition is difficult and expert supervision is expensive.

Few-Shot Learning (FSL) aims to mitigate this limitation by enabling models to generalize from only a handful of labeled examples through the use of prior knowledge \cite{fei-feiOneShotLearningObject2006}. Instead of requiring large training datasets, FSL methods learn transferable representations that allow rapid adaptation to new classes or tasks. This paradigm has demonstrated impressive results across several areas, including image classification \cite{wangGeneralizingFewExamples2020}, object detection \cite{zhangGraphInformationAggregation2024}, and reinforcement learning \cite{wangReinforcementLearningFewShot2023}.

Few-Shot Semantic Segmentation (FSS) extends FSL to dense prediction tasks \cite{shabanOneShotLearningSemantic2017}. In this setting, models are trained through \textit{episodic sampling}: each episode consists of a support set of a few annotated images and a query set of unannotated images. The model learns to segment the query based solely on the support examples, enabling generalization to unseen classes rather than relying on a fixed label space. However, despite rapid progress \cite{wangPANetFewShotImage2019, liAdaptivePrototypeLearning2021, yangMiningLatentClasses2021, zhangFewShotSegmentationCycleConsistent2021}, FSS models exhibit substantial performance degradation when applied across domains with different data distributions. Most works implicitly assume that training and test data come from the same domain, which limits their practical relevance in real-world scenarios where distribution shifts are the norm.

To overcome this limitation, Cross-Domain Few-Shot Semantic Segmentation (CD-FSS) considers a more challenging and realistic setup: source and target domains have disjoint label spaces, and no target-domain samples are available during training \cite{leiCrossDomainFewShotSemantic2022}. CD-FSS therefore addresses both the scarcity of annotations and the domain shift simultaneously.

\begin{figure}[t]
    \centering
    \includegraphics[width=\textwidth]{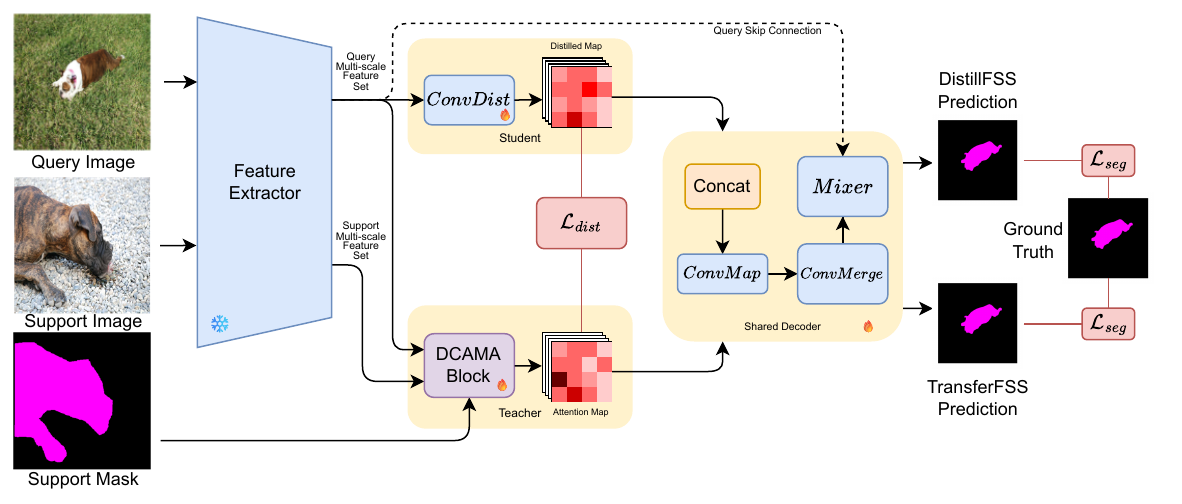}
    \caption{The proposed DistillFSS framework consists of two main components: a student network and a teacher network. The student network is trained on the output of the DCAMA attention block (used as the base model), while the teacher network is designed to focus on the support set. Although the student network has no direct access to the support set, it implicitly encodes its information through the distillation process. The ice icon denotes frozen layers, while the fire icon indicates trainable layers.}
    \label{fig:framework}
\end{figure}

Existing CD-FSS methods primarily aim to transform domain-specific features into domain-agnostic representations or to establish pixel-level correlations between support and query images \cite{leiCrossDomainFewShotSemantic2022}. However, several limitations persist. First, many methods show limited generalization to novel domains, especially when distribution shifts are substantial. Second, most approaches adopt a uniform strategy across domains, overlooking the potential benefits of domain-adaptive mechanisms. Third, standard evaluation settings---where each query image is paired with a support sample from the same class---do not fully reflect real-world conditions, in which class labels are unknown and must be inferred automatically. Finally, scalability remains a significant challenge: dense support--query comparisons lead to rapid increases in computational cost when the number of support images grows beyond 1--10.

Test-time Self-FineTuning (TSF) \cite{leiCrossDomainFewShotSemantic2022, chenCrossDomainFewShotSemantic2024} offers partial mitigation by updating the model using the support set during inference. However, even a single forward/backward update introduces non-negligible overhead, and multiple iterations are often required for satisfactory adaptation.

In this paper, we introduce DistillFSS (Fig.~\ref{fig:framework}), a CD-FSS framework that embeds support-set knowledge directly into a student model to address domain shift, scalability, and the limitations of TSF. The method uses a teacher--student architecture \cite{gouKnowledgeDistillationSurvey2021} in which the teacher processes the support set and distills its information into a dedicated layer of the student network. As a result, the student \textit{internalizes} few-shot reasoning during training and no longer requires episodic support at inference. Combined with fine-tuning, DistillFSS produces a lightweight, deployable network capable of fast and scalable inference, while maintaining the ability—at the framework level—to adapt to novel classes from unseen domains by distilling the teacher into new students

We also introduce a new, comprehensive CD-FSS benchmark that integrates datasets from diverse domains---including medical imaging, industrial inspection, and remote sensing---with disjoint label spaces and varying support sizes, offering a realistic and challenging evaluation protocol.

The main contributions of this work are:
\begin{itemize}
\item We propose DistillFSS, a CD-FSS framework that synthesizes support-set knowledge into the student model via knowledge distillation, enabling efficient and deployable few-shot segmentation.
\item We introduce a new benchmark spanning multiple domains and varying support-image configurations, designed to evaluate CD-FSS methods realistically.
\item We show that DistillFSS achieves competitive or superior performance while significantly reducing computational cost and improving scalability, making it suitable for real-world deployment.
\end{itemize}

The remainder of the paper is organized as follows. Section \ref{sec:related} reviews related work. Section \ref{sec:method} presents the proposed DistillFSS framework. Section \ref{sec:experiments} reports the experimental results, and Section \ref{sec:conclusion} concludes the paper.
\section{Related Work}
\label{sec:related}

\subsection{Few-Shot Semantic Segmentation}

Few-Shot Semantic Segmentation aims to generate pixel-wise predictions for novel classes using only a few labeled support images. In the FSS paradigm, models are trained via episodic sampling: each episode consists of a support set with a few annotated examples and a query set with unannotated images. During meta-training, classes are drawn from a pool of \textit{base classes}, and the model learns to segment query images conditioned on the information provided by the support set. This enables generalization to \textit{novel classes} not encountered during training~\cite{shabanOneShotLearningSemantic2017}.

Existing methodologies can be broadly grouped into \textit{prototype}-based and \textit{affinity}-based approaches. Prototype-based methods \cite{wangPANetFewShotImage2019, yangMiningLatentClasses2021, zhangSGOneSimilarityGuidance2020, dingSelfregularizedPrototypicalNetwork2023,fatehMSDNetMultiscaleDecoder2025,fatehMSDNetMultiscaleDecoder2025} represent support images as class prototypes using masked average pooling and compare them with query features using non-parametric similarity measures. Although effective, these models compress spatially rich feature maps into global prototype vectors, which leads to substantial loss of spatial information \cite{liAdaptivePrototypeLearning2021, zhangFewShotSegmentationCycleConsistent2021}. To mitigate this issue, several works employ clustering or expectation-maximization to derive multiple prototypes corresponding to different object parts \cite{yangPrototypeMixtureModels2020, zhangPyramidGraphNetworks2019}.

Affinity-based methods \cite{zhangFewShotSegmentationCycleConsistent2021, minHypercorrelationSqueezeFewShot, langLearningWhatNot2022, pengHierarchicalDenseCorrelation2023, shiDenseCrossQueryandSupportAttention2022, huLearningForegroundInformation2024, jinEnhancedApproachFewshot2024, zhangEfficientSamplingbasedGaussian2025} instead establish dense pixel-to-pixel correspondences between support and query features. These approaches capture fine-grained spatial information, but their reliance on dense correlation maps makes them computationally demanding and limits their scalability in practical deployments.

Both prototype-based and affinity-based FSS methods remain sensitive to domain gaps between source and target data, resulting in degraded performance in cross-domain settings. Moreover, most FSS approaches are tailored to binary segmentation and handle multiclass scenarios via one-vs-all strategies that treat classes independently. Only recently, a dedicated multiclass FSS framework has been introduced \cite{demarinisLabelAnythingMultiClass2025}.

\subsection{Cross-Domain Semantic Segmentation}

Cross-domain semantic segmentation encompasses two primary subfields: Domain Adaptive Semantic Segmentation (DASS) and Domain Generalized Semantic Segmentation (DGSS).

DASS leverages both source-domain data and labeled or unlabeled target-domain data to improve generalization to the target distribution \cite{lvCrossDomainSemanticSegmentation2020, liuUndoingDamageLabel2022}. Recent unsupervised DASS methods include adversarial approaches \cite{hoffmanCyCADACycleConsistentAdversarial2018, longConditionalAdversarialDomain2018}, which align feature distributions using GANs or Fourier-based transformations, and self-training approaches \cite{zouUnsupervisedDomainAdaptation2018}, which refine predictions using pseudo-labels generated from the target domain.

DGSS instead focuses on enhancing robustness to domain shift without access to target-domain data. Approaches in this area include Normalization and Whitening (NW) techniques \cite{panSwitchableWhiteningDeep2019, pengSemanticAwareDomainGeneralized2022}, which normalize or whiten feature statistics across domains, and Domain Randomization (DR) methods \cite{pengGlobalLocalTexture2021, huangFSDRFrequencySpace2021}, which generate diversified versions of source images to improve generalization.

Unlike traditional DASS and DGSS, Cross-Domain Few-Shot Semantic Segmentation operates under stricter constraints: target-domain images are unavailable during training, and label spaces between domains are disjoint \cite{chenCrossDomainFewShotSemantic2024}. These conditions make CD-FSS substantially more challenging.

\subsection{Cross-Domain Few-Shot Semantic Segmentation}

CD-FSS extends FSS to scenarios where source and target domains differ in both feature distributions and label spaces. This introduces a combination of few-shot supervision and domain shift, making the problem significantly more complex than standard FSS.

Several recent works attempt to address this challenge. PixDA \cite{taveraPixelbyPixelCrossDomainAlignment2022} reduces domain gaps via pixel-level adversarial alignment. RTD \cite{wangRememberDifferenceCrossDomain2022} enhances generalization by transferring intra-domain style representations through a meta-memory mechanism. Lu et al.~\cite{luCrossdomainFewshotSegmentation2022} propose transductive fine-tuning strategies that implicitly supervise query images using support labels. PATNet \cite{leiCrossDomainFewShotSemantic2022} provides a comprehensive evaluation protocol for CD-FSS and converts domain-specific features into domain-agnostic ones. RestNet \cite{huangRestNetBoostingCrossDomain2023} employs lightweight attention mechanisms and residual feature merging to preserve key information while reducing overfitting. DMTNet \cite{chenCrossDomainFewShotSemantic2024} introduces a self-matching transformation module to map domain-specific query features to domain-agnostic representations and uses a dual hypercorrelation module with Test-time Self-FineTuning to enhance segmentation performance.

Despite these advances, existing CD-FSS approaches remain constrained by three significant limitations: (\textit{i}) limited generalization to unseen domains; (\textit{ii}) heavy reliance on dense pixel-to-pixel comparisons, leading to high computational cost; and (\textit{iii}) poor scalability as the number of support images increases. 

In contrast, our proposed DistillFSS shifts the focus from explicit support--query matching to embedding support-set knowledge directly into the model's parameters via knowledge distillation. By transferring support information to a dedicated layer in the student model, DistillFSS eliminates pixel-level matching during inference, resulting in a lightweight, scalable, and deployable framework that efficiently adapts to novel domains.
\section{Method}
\label{sec:method}

\subsection{Problem Formulation}

Given a \textit{source domain} \( (\mathcal{X}_s, \mathcal{Y}_s) \) and a \textit{target domain} \( (\mathcal{X}_t, \mathcal{Y}_t) \), where \( \mathcal{X}_s \neq \mathcal{X}_t \) (input domain shift) and \( \mathcal{Y}_s \cap \mathcal{Y}_t = \emptyset \) (disjoint label spaces), Cross-Domain Few-Shot Semantic Segmentation seeks to train a model \( M_\theta \) able to segment a query image \( \mathcal{Q} \in \mathcal{X}_t \) using a \textit{support set} \( \mathcal{S} \). Traditionally, the support set consists of \( N \) classes, each represented by \( K \) support images with corresponding binary masks:
\begin{equation}
\mathcal{S} = \{ (\mathcal{I}_i, \mathcal{M}_i) \}_{i=1}^{N \times K},
\end{equation}
where \( \mathcal{I}_i \) and \( \mathcal{M}_i \) belong to \( (\mathcal{X}_s, \mathcal{Y}_s) \) during training and to \( (\mathcal{X}_t, \mathcal{Y}_t) \) during testing.

Following the formulation in \cite{demarinisLabelAnythingMultiClass2025}, we relax the standard \( N \times K \) setting by associating each support image \( \mathcal{I}_m \) with a \textit{multiclass mask} \( \mathcal{M}_m \) that encodes all classes present in the image, i.e., \( \mathcal{M}_m[i,j] \in \{0,\ldots,N\} \). The support set thus becomes:
\begin{equation}
\mathcal{S} = \{ (\mathcal{I}_m, \mathcal{M}_m) \}_{m=1}^{M},
\end{equation}
where \( M \) is the number of support images, not necessarily equal to \( N \times K \). This formulation is particularly advantageous when support images contain multiple classes, as it allows defining the entire support set with only \( M \) shots while implicitly modeling all relevant categories.

\subsection{Method Overview}

The goal of DistillFSS (Fig.~\ref{fig:framework}) is to eliminate the need for support images at inference while enabling the framework to be extended to novel classes across unseen domains through additional teacher--student distillation cycles. To achieve this, DistillFSS adopts a two-stage teacher--student design, following standard knowledge-distillation paradigms (e.g., \cite{gouKnowledgeDistillationSurvey2021, hintonDistillingKnowledgeNeural2015}), in which the teacher encodes support-set information and transfers it to a specialized student model.

First, we adopt a few-shot segmentation backbone based on dense cross-attention (DCAMA \cite{shiDenseCrossQueryandSupportAttention2022}) as the \textit{teacher}. Given both the query and the support images, the teacher produces attention maps that guide the segmentation. The teacher is fine-tuned on the support set of the target domain, a process we refer to as \textit{TransferFSS}, so that the attention maps and their weights are adapted to the new domain.

Second, we introduce a \textit{student} network that receives only the query features. A lightweight convolutional block, termed \textit{ConvDist}, is trained to mimic the teacher's attention maps. The student and teacher share the same decoder and are supervised with a segmentation loss. At the same time, the student also minimizes a distillation loss that aligns its output with the teacher's attention maps. After training, the student alone is used for inference, making the model support-free, lightweight, and scalable.

This two-stage process turns explicit support–query interactions into an internal representation learned by the student, thereby combining the adaptivity of few-shot learning with the efficiency required for real-world deployment.

\subsection{Base Model}
\label{sec:base_model}

Our method builds upon the DCAMA architecture introduced in \cite{shiDenseCrossQueryandSupportAttention2022}, a few-shot segmentation framework that leverages dense cross-attention to correlate pixel-level features between query and support images. DCAMA consists of three sequential stages: multi-scale feature extraction, dense cross-attention, and a multi-scale aggregation decoder.

\paragraph{Feature Extraction}

We employ a shared, pre-trained backbone to extract hierarchical feature maps from both query and support images. Instead of relying on a single output layer, we extract a multi-scale set of features \( F = \{F_1, \ldots, F_{|L|}\} \) from selected intermediate layers \( L \). Each feature map \( F_i \) corresponds to a scale \( j \in R = \{1/s^n\}_{n=N_{\min}}^{N_{\max}} \), where \( N_{\min}=2 \) and \( N_{\max}=5 \). High-resolution features at scale \(1/s^{N_{\min}}\) are reserved for skip connections in the decoder.

Let \( F_{q_i} \) and \( F_{s_i} \) denote the query and support feature maps at layer \( i \), with dimensions \( C_j \times H_j \times W_j \). Prior to attention, spatial dimensions are flattened to sequences of size \( N_j \times C_j \), where \( N_j = H_j W_j \), and positional encodings are added. The corresponding binary support mask is reshaped into \( \mathcal{M} \in \mathbb{R}^{N_j \times 1} \).

\paragraph{Dense Cross-Attention}

For each layer, attention maps are computed by projecting query and support features into query–key embeddings:
\begin{align}
Q_i &= F_{q_i} W_j^Q, \\
K_i &= F_{s_i} W_j^K, \\
\mathcal{C}_i &= \text{softmax}\left( \frac{Q_i K_i^\top}{\sqrt{d_k}} \right)\mathcal{M},
\end{align}
where \( W_j^Q \) and \( W_j^K \) are learnable projection matrices at scale \( j \), and \( d_k \) is the channel dimension. The resulting attention map \( \mathcal{C}_i \) reweights query features based on their similarity to support foreground regions.

\paragraph{Multi-Scale Aggregation}

Attention maps \( \mathcal{C} = \{\mathcal{C}_1, \ldots, \mathcal{C}_{|L|}\} \) are grouped by scale and concatenated to obtain tensors of size \( L_j \times H_j \times W_j \). The decoder then processes these tensors through three modules:

\begin{enumerate}
\item \textit{ConvMapper}: Projects concatenated attention maps into a richer feature space using three convolutional layers per scale.
\item \textit{ConvMerge}: Fuses features across scales by iteratively upsampling and merging representations from coarser to finer resolutions.
\item \textit{Mixer}: Combines the aggregated attention features with high-resolution query features via skip connections, refining the representation through three convolutional blocks to produce the final segmentation map.
\end{enumerate}

\subsection{TransferFSS}

We adopt DCAMA as our base model due to its simplicity and effectiveness, which make it suitable for transfer learning and distillation. Before training, we introduced an architectural refinement: skip connections concatenate only query features, excluding support features. Since query and support images lack spatial alignment, mixing their features is uninformative and complicates support-independent inference. We observed improved performance after retraining the model with this modification.

\paragraph{Target-Domain Adaptation}

Given the optimized DCAMA, we fine-tune the \textit{ConvMapper} block using the target-domain support set, treating it as training data. In each epoch, one support image serves as the query, and \(R\) randomly sampled support images serve as conditioning inputs. We refer to this fine-tuning stage as \textit{TransferFSS}.

The motivation is that the \textit{ConvMapper} block, originally trained on the source domain, may overfit to source-specific characteristics, leading to suboptimal feature weighting in unseen domains. Even when the final segmentation mask is inaccurate, specific attention maps \( \mathcal{C}_i \) may still be reliable. Since \textit{ConvMapper} determines how attention maps contribute to the final prediction, adjusting its parameters is crucial for adapting to a new domain.

\begin{figure}[t]
    \centering
    \input{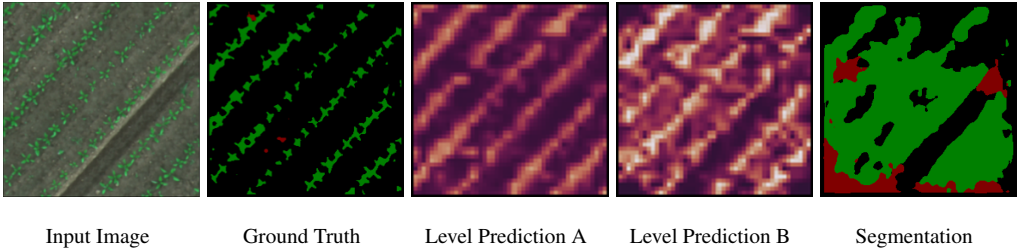}
    \caption{Example of a DCAMA prediction on the WeedMap dataset, showing two attention maps for the crop class taken from different depths in the network. Level A comes from an earlier stage (low-level features), while level B corresponds to a later stage (high-level features). Although level A is more accurate, the model assigns higher importance to level B, leading to an incorrect final prediction.}
    \label{fig:preliminary}
\end{figure}

Figure \ref{fig:preliminary} highlights this behavior by comparing attention maps at two different depths in the network. Level A corresponds to an early stage dominated by low-level cues, while Level B originates from a later stage with predominantly high-level semantics. Although Level A aligns more closely with the ground truth, the model disproportionately relies on Level B, which leads to suboptimal segmentation. This discrepancy also suggests that only a subset of levels contributes meaningfully to the prediction, implying that the model uses more depth than it actually needs and that a more compact architecture could capture the relevant information just as effectively.

\subsection{DistillFSS}

After fine-tuning, the model may eventually overfit to the specific support set, suggesting that it may be unnecessary at inference time if the support set does not change. Removing support--query interactions would significantly reduce computational cost.

\paragraph{Distilled Architecture}

In DCAMA, the support set is used exclusively within the attention blocks to compute attention maps \( \mathcal{C}_i \). Within DistillFSS, the teacher and the student share the same backbone and decoder inherited from the base model; the only modules that differ are the support-dependent attention blocks, which the student replaces with \textit{ConvDist}. Our key idea is to distill the behavior of these blocks into a lightweight module that operates solely on the query image. We denote the resulting model as \textit{DistillFSS}.

To replace each attention block, we introduce \textit{ConvDist}, a compact convolutional module:
\begin{equation}
    ConvDist = \text{Sigmoid} \circ Conv_{1 \times 1} \circ \text{ReLU} \circ Conv_{3 \times 3},
\end{equation}
where each convolution preserves the input channel dimensionality. A separate \textit{ConvDist} module is distilled for each scale.

\paragraph{Optimization}

Distillation is performed on the support set over \( E \) epochs. We combine fine-tuning and distillation through the composite loss:
\begin{equation}
L = L_{dist} + L_{seg}(out_{Student}) + L_{seg}(out_{Teacher}),
\end{equation}
where both teacher and student outputs contribute to segmentation supervision.

For segmentation, we adopt the Focal Loss \cite{linFocalLossDense2017}:
\begin{equation}
L_{seg}(p_t) = -\alpha_t (1 - p_t)^\gamma \log(p_t).
\end{equation}

For distillation, we avoid Cross-Entropy due to dropout-induced deviations from valid probability distributions and instead employ Mean Squared Error:
\begin{equation}
L_{dist} = \frac{1}{N}\sum_{i=1}^{N} \text{MSE}\big(\mathcal{C}_i, ConvDist(F_{q_i})\big),
\end{equation}
where \( \mathcal{C}_i \) are the teacher attention maps and \( N \) the number of scales. This training strategy yields a lightweight model that approximates the behavior of DCAMA's attention blocks while eliminating support-set usage during inference. 

\begin{figure}[t]
    \centering
    \includegraphics[width=\textwidth]{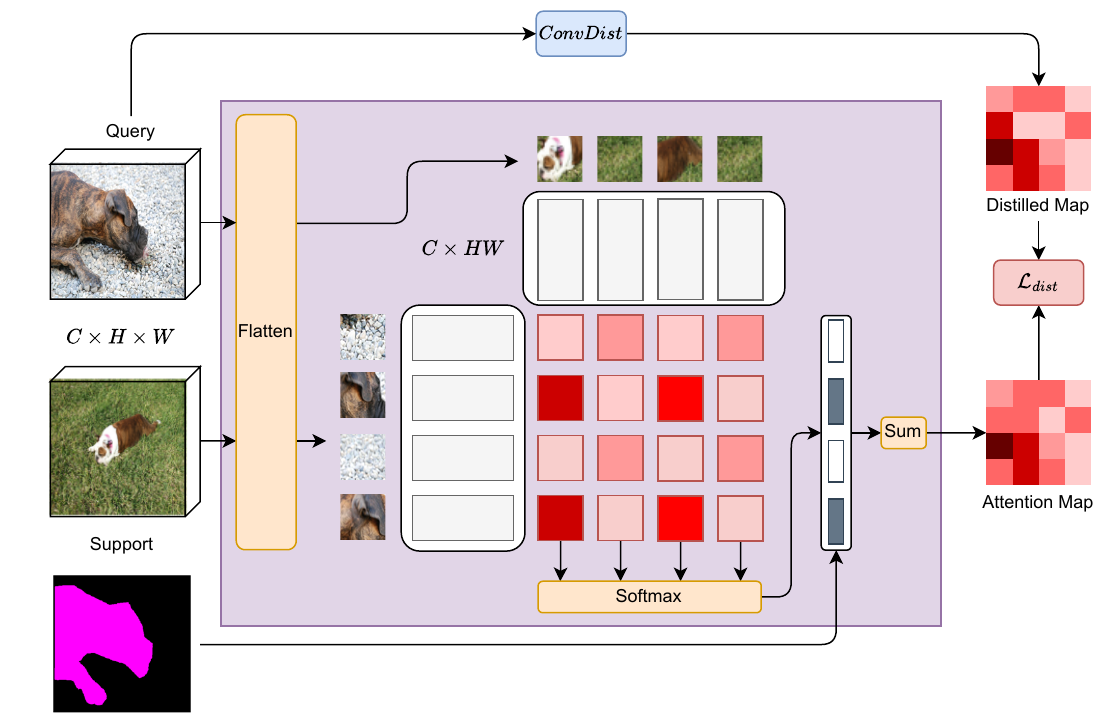}
    \caption{Overview of the distillation process in DistillFSS for a single scale. A DCAMA attention block is distilled into a lightweight \textit{ConvDist} module, which replicates its behavior without requiring support images at inference. Knowledge transfer is enforced through the distillation loss \( L_{dist} \).}
    \label{fig:distillation}
\end{figure}

Figure~\ref{fig:distillation} illustrates the internal process of distilling a DCAMA attention block. The teacher computes attention by correlating flattened query and support features, applying a softmax over the support positions, and weighting the result with the support mask. \textit{ConvDist}, instead, receives only the query feature map and predicts a distilled map in the same spatial format. The two maps are aligned through the distillation loss.

\paragraph{Generalization}

The proposed distillation idea is general and can be applied to any few-shot model. Given a model \( M_\theta \) with modules \( \mathcal{B}=\{\mathcal{B}_1,\ldots,\mathcal{B}_B\} \), we identify the subset \( \mathcal{B}_\mathcal{S} \subset \mathcal{B} \) that depends on the support set. For each \( \mathcal{B}_i \in \mathcal{B}_\mathcal{S} \), we may learn a corresponding simplified module \( \mathcal{T}_i' \) that does not require support-set inputs. While we found that a convolutional block is suitable for replacing DCAMA attention modules, other few-shot architectures may require different distillation modules. A full exploration of this generalization is beyond the scope of this work.

\subsection{Implementation Details}

We train the base model using AdamW with a learning rate of \(1\mathrm{e}{-3}\). Training is stopped early once the mIoU on the support set stops improving. Experiments employ two feature extractors: ResNet-50 \cite{heDeepResidualLearning2016a} and Swin-B \cite{liuSwinTransformerHierarchical2021}. For the Focal Loss, we set \( \gamma = 2.0 \) and \( \alpha_{t} = 1.0 \), i.e., no class reweighting is applied.
\section{Experiments}
\label{sec:experiments}

\subsection{Setting}

We evaluated our method and all baselines using the mean Intersection over Union (mIoU), defined as:
\begin{equation}
    \text{mIoU} = \frac{1}{N} \sum_{i=1}^{N} \frac{TP_i}{TP_i + FP_i + FN_i},
\end{equation}
where \(N\) is the number of classes and \(TP_i\), \(FP_i\), and \(FN_i\) denote true positives, false positives, and false negatives for class \(i\).

We first compared our optimized architecture (removing support-set skip connections) with the original DCAMA on fold~0 of the Pascal-$5^i$ dataset \cite{shabanOneShotLearningSemantic2017} in a 1-way 1-shot episodic setting. The optimized version achieved 65.94\% mIoU, while the original obtained 62.62\%. This confirms that support skip connections are unnecessary in DCAMA. The optimized model is therefore used as the base model for all subsequent experiments.

To evaluate cross-domain performance, we established a new CD-FSS benchmark incorporating seven challenging datasets, each with distinct characteristics:

\begin{itemize}
    \item Lung Nodule \cite{namLungCancerSegmentation2024}.  
    CT-based lung nodule segmentation, where nodules are small and visually similar to surrounding tissue, replacing the simpler Chest X-ray dataset \cite{LungSegmentationChest} commonly used in prior work.

    \item ISIC 2018 \cite{codellaSkinLesionAnalysis2019}.  
    Skin lesion segmentation with three classes (melanoma, nevus, seborrheic keratosis), characterized by subtle inter-class variations and acquisition noise.

    \item KVASIR-Seg \cite{jha2020kvasir}.  
    Polyp segmentation in colonoscopy images, where polyps frequently blend into surrounding tissue.

    \item Nucleus \cite{longMicroscopyCellNuclei2020}.  
    Nucleus segmentation in histopathology involves significant variability in cell and background appearances.

    \item WeedMap \cite{saWeedMapLargeScaleSemantic2018}.  
    Remote-sensing weed segmentation is affected by strong class imbalance and visual similarity between weeds and crops.

    \item Pothole-mix \cite{ranieriPotholeMix2022}.  
    Aggregated pothole segmentation datasets exhibiting significant variation in illumination and scene geometry.

    \item Industrial-$5^i$ \cite{shiFewshotSemanticSegmentation2023}.  
    Industrial defect segmentation with 20 classes across four folds, containing minor, heterogeneous defects.
\end{itemize}

When available, we used the official train/test splits; otherwise, 20\% of the dataset was sampled as the test set. Support images were drawn from the training partitions, and all test images were used as queries.

We adopted a more realistic evaluation setting than standard episodic testing \cite{chenCrossDomainFewShotSemantic2024, leiCrossDomainFewShotSemantic2022}. Traditional CD-FSS benchmarks sample a random class paired with a single support and query image. This is unrealistic because: (\textit{i}) the query class is unknown at inference; (\textit{ii}) the support set in real deployments is fixed and known in advance; (\textit{iii}) support and queries typically originate from different domains. In our evaluation, the support set is therefore fixed, contains samples from all classes, and is drawn from a different domain than the queries. This better reflects real-world cross-domain settings.

For 1-class datasets (and WeedMap), we used \(K=M \in \{5,10,25,50\}\). For ISIC (three classes), \(K = 3,5,10,20\) and thus \(M=9,15,30,60\). For Industrial-$5^i$ (five classes), \(K = 2,4,8,16\) and \(M=10,20,40,80\). For Industrial-$5^i$, the reported results are averaged over the four folds. Higher-shot configurations always include the support samples of lower-shot ones.

\subsection{Comparison with State-of-the-Art Methods}

\begin{table}[h]
    \centering
    \tiny
    \setlength{\tabcolsep}{0pt}

    % Define a fixed width for the first two columns combined to ensure alignment
    % We use 'p' columns for fixed width, but we suppress the extra padding with @{}
    \newcommand{\firstcolwidth}{1.2cm}
    \newcommand{\secondcolwidth}{1.0cm}

    % --- TOP TABLE ---
    \begin{tabular*}{\textwidth}{@{} p{\firstcolwidth} @{\hspace{1em}} p{\secondcolwidth} @{\extracolsep{\fill}} rrrrrrrrrrrrrrrr }
        \toprule \multicolumn{2}{c}{}                                          & \multicolumn{4}{c}{Lung Nodule} & \multicolumn{4}{c}{ISIC} & \multicolumn{4}{c}{KVASIR-Seg} & \multicolumn{4}{c}{Nucleus} \\
        \cmidrule{3-6} \cmidrule{7-10} \cmidrule{11-14} \cmidrule{15-18} Model & Backbone                        & 5                        & 10                             & 25                         & 50             & 9              & 15             & 30             & 60             & 5              & 10             & 25             & 50             & 5              & 10             & 25             & 50             \\
        \midrule BAM                                                           & ResNet-50                       & 0.17                     & 0.17                           & 0.20                       & 0.19           & 9.67           & 8.34           & 8.47           & 8.69           & 18.96          & 19.28          & 23.05          & 23.03          & 11.03          & 11.32          & 12.07          & 11.05          \\
        HDMNet                                                                 & ResNet-50                       & 0.04                     & 0.12                           & 0.14                       & 0.16           & 9.16           & 7.82           & 8.20           & 7.77           & 28.97          & 29.70          & 34.13          & 33.37          & 19.64          & 19.92          & 21.17          & 21.77          \\
        LabelAnything                                                          & ViT-B                           & 0.05                     & 0.04                           & 0.04                       & 0.04           & 7.20           & 7.65           & 11.39          & 13.90          & 27.76          & 27.78          & 27.75          & 27.67          & 19.99          & 19.46          & 18.64          & 18.46          \\
        PATNet                                                                 & ResNet-50                       & 0.08                     & 0.22                           & 0.27                       & 0.30           & \textbf{16.01} & 13.46          & 15.95          & 16.97          & 41.94          & 44.93          & 49.72          & 46.78          & 37.74          & 32.01          & 33.38          & 33.43          \\
        DMTNet                                                                 & ResNet-50                       & 0.16                     & 0.16                           & 0.16                       & 0.16           & 15.00          & 15.07          & 15.55          & 17.35          & \textbf{47.78} & 48.98          & 50.01          & 46.54          & 19.24          & 19.76          & 21.50          & 23.53          \\
        DCAMA                                                                  & ResNet-50                       & 0.11                     & 0.00                           & 0.23                       & 0.24           & 3.36           & 8.80           & 6.97           & 7.59           & 36.08          & 33.40          & 34.34          & 29.82          & 19.88          & 23.21          & 19.41          & 19.70          \\
        DCAMA                                                                  & Swin-B                          & 0.18                     & 0.27                           & 0.21                       & 0.18           & 8.67           & 7.25           & 7.22           & 7.72           & 29.42          & 28.65          & 28.37          & 28.10          & 28.82          & 24.78          & 31.80          & 28.64          \\
        TransferFSS                                                            & ResNet-50                       & 8.51                     & \textbf{14.74}                 & \textbf{22.24}             & \textbf{24.28} & 9.38           & 15.65          & 14.34          & 15.00          & 31.99          & 41.18          & 45.72          & 48.58          & 19.09          & 23.28          & 21.96          & 22.71          \\
        TransferFSS                                                            & Swin-B                          & 3.43                     & 2.29                           & 1.54                       & 2.51           & 14.35          & 15.81          & \textbf{20.48} & 22.15          & 45.18          & 49.28          & 61.84          & \textbf{59.97} & \textbf{73.12} & \textbf{79.01} & \textbf{69.47} & 79.39          \\
        DistillFSS                                                             & ResNet-50                       & \textbf{8.91}            & 11.70                          & 16.59                      & 20.02          & 11.24          & 12.48          & 12.36          & 12.22          & 29.75          & 42.72          & 44.90          & 46.18          & 12.76          & 20.71          & 25.22          & 21.12          \\
        DistillFSS                                                             & Swin-B                          & 3.31                     & 3.54                           & 3.42                       & 4.87           & 13.31          & \textbf{15.95} & 19.32          & \textbf{23.41} & 37.29          & \textbf{52.83} & \textbf{61.87} & 57.09          & 69.57          & 71.16          & 68.19          & \textbf{79.96} \\
        \bottomrule
    \end{tabular*}

    \vspace{1pt}

    % --- BOTTOM TABLE ---
    % Note: To make the third column start at the EXACT same spot, we need to compensate
    % for the fact that tabular* adds the "fill" space BEFORE the first 'r' column.
    % In the top table, that fill is small (18 cols). In the bottom, it's large (14 cols).
    % The simplest hack: Force the first 'r' column to NOT have the fill immediately before it,
    % or use a fixed spacer.
    %
    % However, the cleanest visual fix without complex calculations is to keep the left side rigid
    % and let the right side float.
    \begin{tabular*}{\textwidth}{@{} p{\firstcolwidth} @{\hspace{1em}} p{\secondcolwidth} @{\extracolsep{\fill}} rrrrrrrrrrrr }
        \multicolumn{2}{c}{}                                  & \multicolumn{4}{c}{WeedMap} & \multicolumn{4}{c}{Pothole-mix} & \multicolumn{4}{c}{Industrial-$5^i$} \\
        \cmidrule{3-6} \cmidrule{7-10} \cmidrule{11-14} Model & Backbone                    & 5                               & 10                                           & 25             & 50             & 5              & 10             & 25             & 50             & 10            & 20             & 40             & 80             \\
        \midrule BAM                                          & ResNet-50                   & 6.63                            & 5.53                                         & 6.36           & 6.16           & 1.46           & 1.53           & 3.90           & 2.23           & \textbf{4.98} & 4.57           & 4.88           & 4.86           \\
        HDMNet                                                & ResNet-50                   & 1.24                            & 3.71                                         & 3.81           & 4.04           & 3.49           & 3.46           & 3.35           & 3.51           & 1.92          & 3.13           & 4.11           & 4.00           \\
        LabelAnything                                         & ViT-B                       & 2.28                            & 2.21                                         & 3.69           & 3.74           & 11.76          & 11.13          & 8.75           & 8.58           & 1.83          & 2.16           & 1.94           & 2.12           \\
        PATNet                                                & ResNet-50                   & 6.55                            & 6.30                                         & 6.95           & 6.96           & 8.85           & 15.35          & 9.91           & 8.80           & 1.61          & 2.06           & 2.33           & 2.14           \\
        DMTNet                                                & ResNet-50                   & 2.61                            & 2.17                                         & 2.40           & 2.30           & 10.69          & 8.43           & 7.81           & 8.16           & 1.26          & 1.34           & 1.81           & 1.86           \\
        DCAMA                                                 & ResNet-50                   & 5.10                            & 5.11                                         & 4.90           & 5.01           & 12.42          & 12.14          & 10.92          & 8.62           & 2.09          & 2.56           & 3.38           & 3.21           \\
        DCAMA                                                 & Swin-B                      & 4.70                            & 6.13                                         & 4.38           & 3.78           & 16.80          & \textbf{18.90} & 11.89          & 10.43          & 0.84          & 1.14           & 1.21           & 1.16           \\
        TransferFSS                                           & ResNet-50                   & 31.73                           & 46.18                                        & 52.09          & 54.03          & 4.04           & 10.19          & 9.60           & 5.34           & 1.53          & 6.52           & 5.27           & 8.76           \\
        TransferFSS                                           & Swin-B                      & \textbf{51.01}                  & \textbf{58.65}                               & 57.55          & \textbf{64.18} & \textbf{17.36} & 15.99          & \textbf{22.66} & 31.77          & 4.09          & \textbf{30.33} & \textbf{42.05} & \textbf{48.19} \\
        DistillFSS                                            & ResNet-50                   & 32.40                           & 47.99                                        & 48.38          & 54.48          & 8.06           & 10.11          & 10.98          & 14.26          & 1.62          & 3.52           & 4.36           & 4.79           \\
        DistillFSS                                            & Swin-B                      & 44.43                           & 55.38                                        & \textbf{59.43} & 61.96          & 17.01          & 17.51          & 14.47          & \textbf{31.96} & 3.50          & 25.48          & 40.86          & 46.09          \\
        \bottomrule
    \end{tabular*}
    \caption{Comparison between DistllFSS and state-of-the-art methods across multiple datasets, reporting mIoU scores varying the number of support images.}
    \label{tab:results}
\end{table}

We compared DistillFSS with state-of-the-art FSS and CD-FSS models. Except for LabelAnything \cite{demarinisLabelAnythingMultiClass2025}, all methods are originally designed for the 1-way setting; we extend them to the \(N\)-way case via one-vs-all. PATNet and DMTNet aggregate predictions across multiple shots via voting, whereas DCAMA and LabelAnything explicitly support multi-shot input. BAM and HDMNet provide fixed-shot weights (1-shot and 5-shot); for larger \(K\), we followed their protocol by repeating the forward pass \(K/5\) times and selecting the highest-confidence prediction.

Since DCAMA’s attention computation scales linearly with the number of shots (key size \(KHW \times HW\)), evaluating large support sets is computationally expensive. We therefore group support images into batches of 10 for DCAMA, TransferFSS, BAM, and HDMNet.

\begin{figure}
  \centering
  \renewcommand{\arraystretch}{0.5}
  \setlength{\tabcolsep}{1pt}
  \begin{tabular}{cccccc}
    % Lung Nodule
    \includegraphics[width=0.14\textwidth]{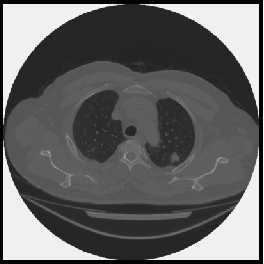} &
    \includegraphics[width=0.14\textwidth]{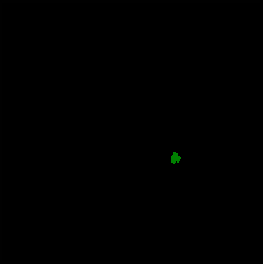} &
    \includegraphics[width=0.14\textwidth]{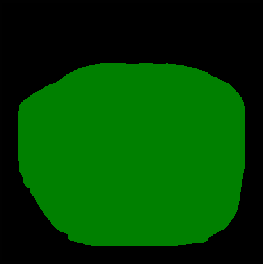} &
    \includegraphics[width=0.14\textwidth]{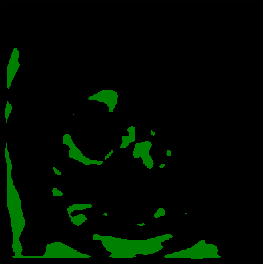} &
    \includegraphics[width=0.14\textwidth]{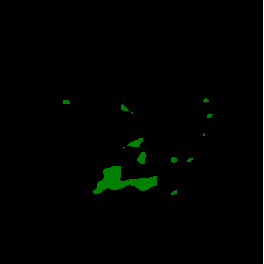} &
    \includegraphics[width=0.14\textwidth]{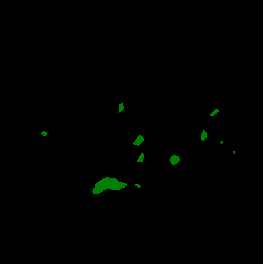} \\
    % ISIC
    \includegraphics[width=0.14\textwidth]{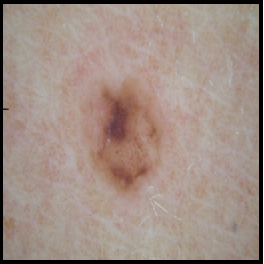} &
    \includegraphics[width=0.14\textwidth]{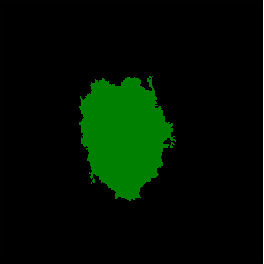} &
    \includegraphics[width=0.14\textwidth]{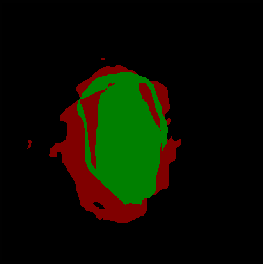} &
    \includegraphics[width=0.14\textwidth]{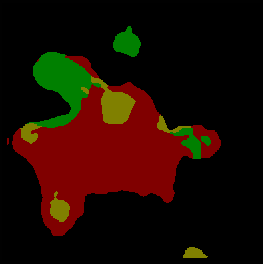} &
    \includegraphics[width=0.14\textwidth]{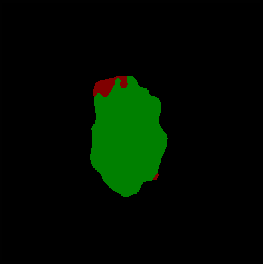} &
    \includegraphics[width=0.14\textwidth]{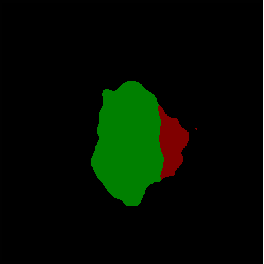} \\
    % KVASIR-Seg
    \includegraphics[width=0.14\textwidth]{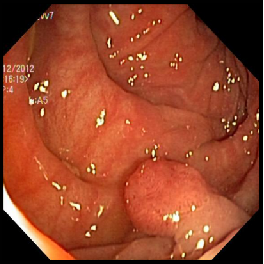} &
    \includegraphics[width=0.14\textwidth]{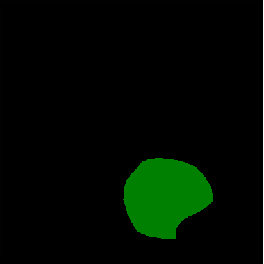} &
    \includegraphics[width=0.14\textwidth]{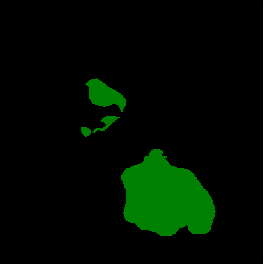} &
    \includegraphics[width=0.14\textwidth]{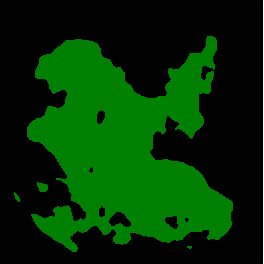} &
    \includegraphics[width=0.14\textwidth]{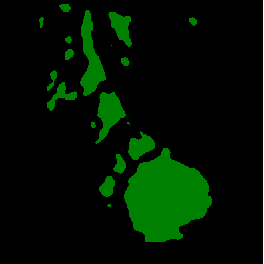} &
    \includegraphics[width=0.14\textwidth]{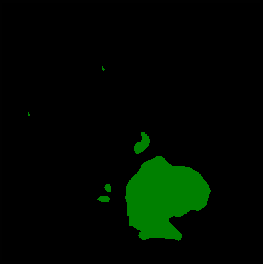} \\
    % Nucleus
    \includegraphics[width=0.14\textwidth]{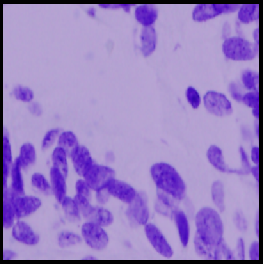} &
    \includegraphics[width=0.14\textwidth]{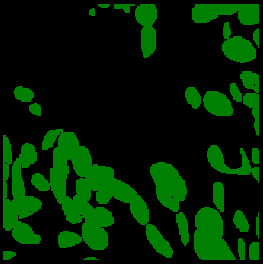} &
    \includegraphics[width=0.14\textwidth]{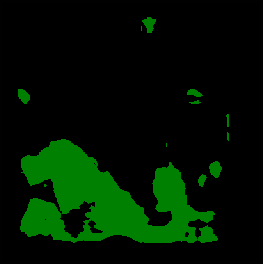} &
    \includegraphics[width=0.14\textwidth]{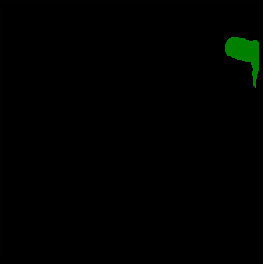} &
    \includegraphics[width=0.14\textwidth]{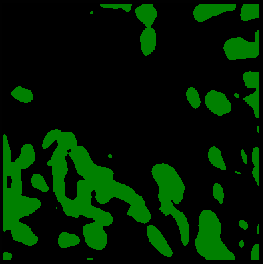} &
    \includegraphics[width=0.14\textwidth]{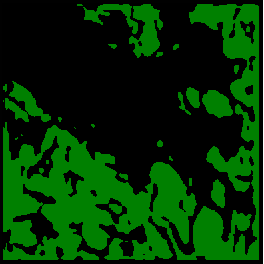} \\
    % WeedMap
    \includegraphics[width=0.14\textwidth]{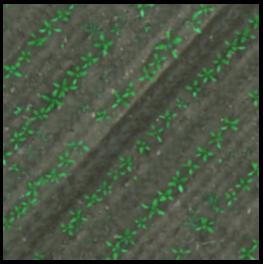} &
    \includegraphics[width=0.14\textwidth]{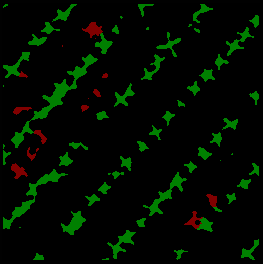} &
    \includegraphics[width=0.14\textwidth]{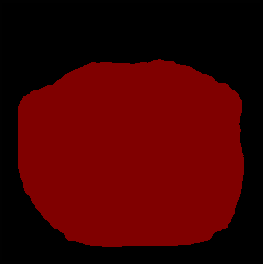} &
    \includegraphics[width=0.14\textwidth]{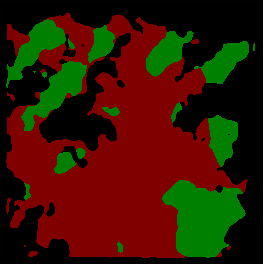} &
    \includegraphics[width=0.14\textwidth]{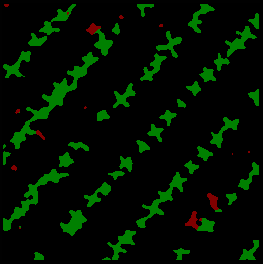} &
    \includegraphics[width=0.14\textwidth]{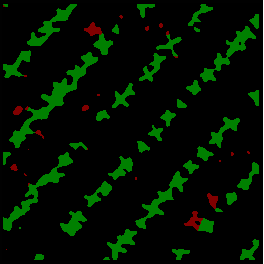} \\
    % Pothole-mix
    \includegraphics[width=0.14\textwidth]{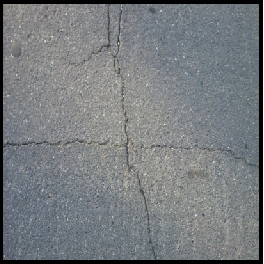} &
    \includegraphics[width=0.14\textwidth]{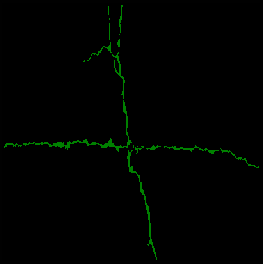} &
    \includegraphics[width=0.14\textwidth]{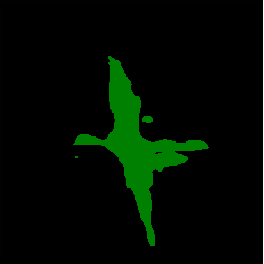} &
    \includegraphics[width=0.14\textwidth]{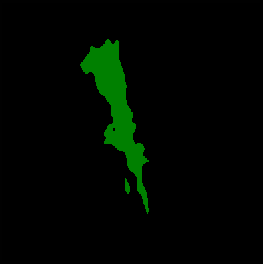} &
    \includegraphics[width=0.14\textwidth]{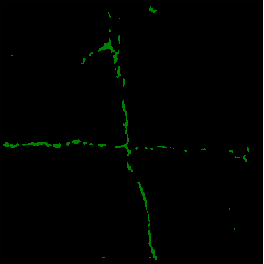} &
    \includegraphics[width=0.14\textwidth]{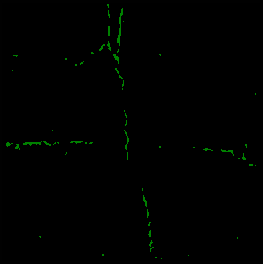} \\
    % Industrial-$5^{i}$
    \includegraphics[width=0.14\textwidth]{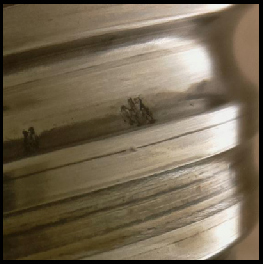} &
    \includegraphics[width=0.14\textwidth]{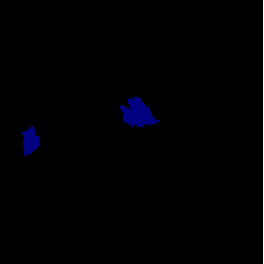} &
    \includegraphics[width=0.14\textwidth]{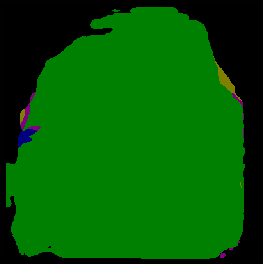} &
    \includegraphics[width=0.14\textwidth]{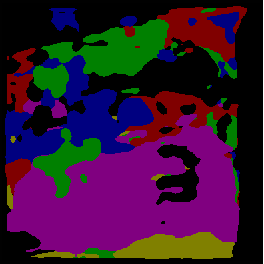} &
    \includegraphics[width=0.14\textwidth]{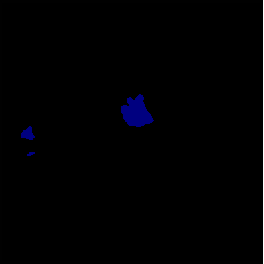} &
    \includegraphics[width=0.14\textwidth]{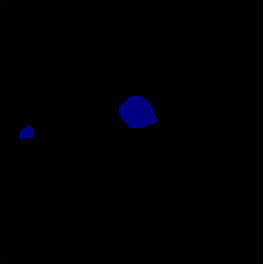} \\
    % Labels
    \scriptsize Query Image & \scriptsize Ground Truth & \scriptsize DMTNet & \scriptsize DCAMA &
    \scriptsize \shortstack{TransferFSS\\(ours)} & \scriptsize \shortstack{DistillFSS\\(ours)}
  \end{tabular}
\caption{Qualitative results across various selected datasets, listed from top to bottom: Lung Nodule, ISIC, KVASIR-Seg, Nucleus, WeedMap, Pothole-mix, and Industrial-$5^{i}$.}
\label{fig:qualit}
\end{figure}

Table~\ref{tab:results} reports all quantitative results. DistillFSS achieves the best or comparable performance across most datasets despite its reduced complexity. TransferFSS and DistillFSS consistently outperform the DCAMA baseline (except on Pothole with \(M=10\)), showing the benefit of fine-tuning and distillation.

The advantage of our method grows with the number of shots. On ISIC, for example, PATNet performs best for \(K=9\) but degrades as \(K\) increases, whereas both TransferFSS and DistillFSS improve. On Industrial-$5^i$, improvements are modest with ResNet-50 but substantial with Swin-B, suggesting that our approach is particularly effective with stronger backbones.

Figure~\ref{fig:qualit} shows qualitative examples. Competing methods struggle in multi-class scenarios, whereas our approach yields sharper and more consistent masks. Both TransferFSS and DistillFSS visibly improve segmentation quality compared to DCAMA.

\subsection{Computational Complexity}

We assessed inference time and peak memory usage across different numbers of shots (\(K=5,10,25,50\)) and ways (\(N=1,3,5\)). All experiments were conducted on an NVIDIA A100 (80~GiB VRAM).

Table~\ref{tab:inference} reports inference times. DistillFSS already matches DCAMA in the 1-way 1-shot case (47~ms vs.~70~ms). As shots and ways increase, the difference becomes dramatic: DCAMA requires 3130~ms in the 5-way 50-shot setting, whereas DistillFSS maintains 66~ms, being independent of the support size. Sequential baselines exhibit increasing inference times as \(K\) and \(N\) grow. LabelAnything is unaffected by \(N\) but slows with large \(K\). DMTNet reaches 32\,024~ms for the 5-way 5-shot setting, making it impractical for deployment.

\begin{table}
\centering
\scriptsize
\setlength{\tabcolsep}{2.5pt} % Reduce padding slightly to fit
\resizebox{\linewidth}{!}{
\begin{tabular}{lrrrrr c rrrrr c rrrrr}
    \toprule
    Ways & \multicolumn{5}{c}{1 Way} && \multicolumn{5}{c}{3 Ways} && \multicolumn{5}{c}{5 Ways} \\
    \cmidrule{2-6} \cmidrule{8-12} \cmidrule{14-18}
    Shots & 1 & 5 & 10 & 25 & 50 && 1 & 5 & 10 & 25 & 50 && 1 & 5 & 10 & 25 & 50 \\
    \midrule
    BAM \cite{langLearningWhatNot2022} & 89 & 171 & 290 & 649 & 1250 && 168 & 413 & 768 & 1855 & 3655 && 242 & 650 & 1251 & 3046 & 6057 \\
    HDMNet \cite{pengHierarchicalDenseCorrelation2023} & 126 & 259 & 473 & 1102 & 2148 && 265 & 680 & 1310 & 3204 & 6352 && 405 & 1101 & 2152 & 5302 & 10562 \\
    LabelAnything \cite{demarinisLabelAnythingMultiClass2025} & 86 & 137 & 196 & 385 & 696 && 86 & 140 & 204 & 402 & 733 && 87 & 143 & 212 & 419 & 765 \\
    PATNet \cite{leiCrossDomainFewShotSemantic2022} & 138 & 190 & 318 & 694 & 1334 && 304 & 452 & 833 & 1982 & 3871 && 458 & 715 & 1347 & 3243 & 6367 \\
    DMTNet \cite{chenCrossDomainFewShotSemantic2024} & 197 & 701 & 1336 & 3256 & 6363 && 472 & 1986 & 3864 & 9498 & 19184 && 742 & 3275 & 6444 & 15859 & 32024 \\
    DCAMA \cite{shiDenseCrossQueryandSupportAttention2022} & 70 & 113 & 174 & 358 & 663 && 111 & 233 & 417 & 967 & 1897 && 149 & 360 & 668 & 1578 & 3130 \\
    DistillFSS (ours) & \textbf{47} & \textbf{67} & \textbf{67} & \textbf{66} & \textbf{66} && \textbf{66} & \textbf{67} & \textbf{67} & \textbf{66} & \textbf{66} && \textbf{66} & \textbf{67} & \textbf{66} & \textbf{66} & \textbf{66} \\
    \bottomrule
\end{tabular}
}
\caption{Comparison of the inference time (ms) of our approach with state-of-the-art methods on different datasets by varying the number of classes (Ways) and support images (Shots).}
\label{tab:inference}
\end{table}

\begin{figure}[t]
    \centering
    \includesvg[width=0.8\textwidth]{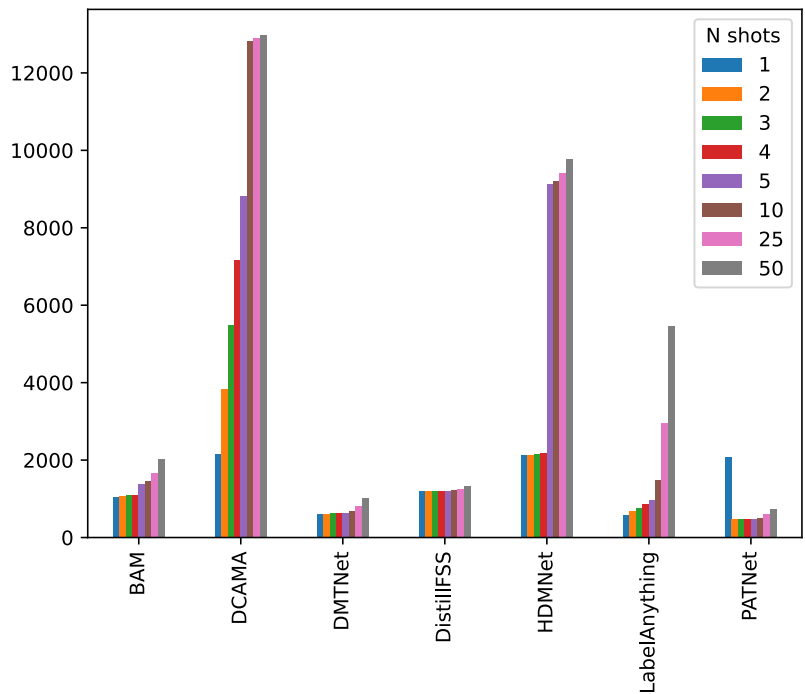}
    \caption{Peak memory consumption (MiB) during the forward pass across models in the 1-way setting.}
    \label{fig:memory}
\end{figure}

Figure~\ref{fig:memory} shows peak memory usage. DistillFSS requires only 1.2~GiB. DCAMA’s memory grows linearly with \(K\) (up to 13~GiB at 10 shots) before stabilizing due to batching. HDMNet shows similar behavior.

\subsection{Ablation Study}
\label{sec:ablation}

We performed ablation studies to evaluate the effect of unfreezing specific components during fine-tuning. As noted in Section~\ref{sec:method}, \textit{ConvMapper} is the minimal module eligible for adaptation. While fine-tuning it improves recognition, the resulting segmentations are coarse, likely due to suboptimal skip connections. We therefore also fine-tuned \textit{ConvSkip} and the \textit{Classifier}. Table~\ref{tab:ablation} reports these results.

\begin{figure*}[t]
    \centering
    \includesvg[width=\linewidth]{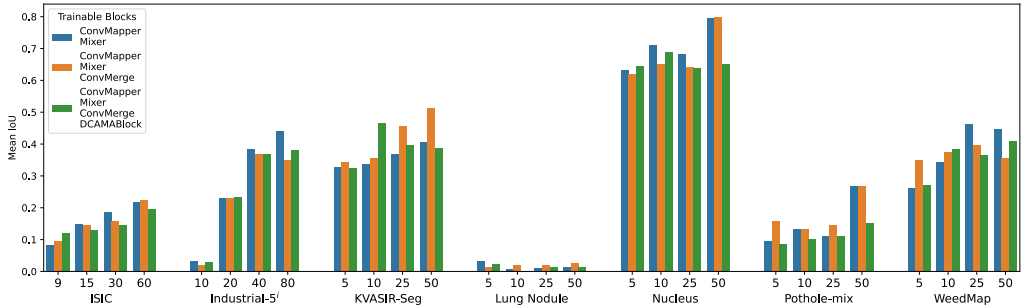}
    \caption{Ablation study on the unfrozen training blocks of DistillFSS across datasets and shot configurations.}
    \label{fig:ablation}
\end{figure*}

We further investigated unfreezing \textit{ConvMerge} and the attention weights \(W_j\). Across datasets, we observed no consistent improvements, indicating that their usefulness is dataset-dependent (Fig.~\ref{fig:ablation}).

Finally, we assessed the contribution of the distillation loss \(L_{dist}\) by training DistillFSS without it. Results in Table~\ref{tab:ablation_dist} show that \(L_{dist}\) is essential: without this loss, the student model fails to replicate the behavior of DCAMA’s attention blocks, leading to significant performance drops.

\begin{table}[t]
    \centering
    \begin{tabular}{crrrr}
    \toprule
    Mixer &     5 &    10 &    25 &    50 \\
    \midrule
    \ding{55}     & 15.75 & 14.43 & 18.55 & 19.94 \\
    \ding{51}     & 31.59 & 49.88 & 56.81 & 64.18 \\
    \bottomrule
\end{tabular}
    \caption{Ablation study on fine-tuning the Mixer Block. Results are reported as mIoU on WeedMap for different shot configurations.}
    \label{tab:ablation}
\end{table}

\begin{table}[t]
    \centering
    \begin{tabular}{crrrr}
    \toprule
    Dist.~loss      &     5 &    10 &    25 &    50 \\
    \midrule
     \ding{55} & 44.43 & 29.4  & 36.16 & 34.54 \\
     \ding{51} & 36.46 & 55.38 & 59.43 & 61.96 \\
\bottomrule
\end{tabular}
    \caption{Ablation study on the distillation loss. Results are reported as mIoU on WeedMap for different shot configurations.}
    \label{tab:ablation_dist}
\end{table}
\section{Conclusion}
\label{sec:conclusion}

This work introduced DistillFSS, a framework for Cross-Domain Few-Shot Semantic Segmentation that replaces the standard few-shot inference pipeline with a model that has already absorbed the functional role of the support set during training. As a result, the method no longer requires support images at test time and avoids the computational cost of pixel-level matching.

Experiments on heterogeneous domains show that this strategy attains competitive or superior performance compared to existing FSS and CD-FSS approaches while remaining efficient and scalable. The advantages become more pronounced as the support set increases, where traditional methods incur rapidly growing costs, making DistillFSS suitable for deployment scenarios that demand fast adaptation under limited resources.

We also introduced a new CD-FSS benchmark covering multiple domains and disjoint label spaces, providing a more realistic evaluation setting than conventional few-shot protocols. Additional analyses identified the model components most sensitive to specialization and clarified the contribution of the optimization strategy to the final accuracy.

Overall, the results indicate that rethinking the role of the support set—shifting its contribution from inference to adaption—offers a promising direction for improving cross-domain adaptation in few-shot segmentation. Future work could explore how this perspective transfers to different architectures, how to refine specialization for multi-class settings, and how to assess domain complexity to estimate the required support-set diversity.

% \appendix
% \section{Example Appendix Section}




\section*{Acknowledgment}

We acknowledge the CINECA award under the ISCRA initiative, which gave us access to computing resources and support. The research of  Pasquale De Marinis is funded by a Ph.D.~fellowship within the framework of the Italian ``D.M.~n.~352, April 9, 2022'' -- under the National Recovery and Resilience Plan, Mission 4, Component 2, Investment 3.3 -- Ph.D.~Project ``Computer Vision techniques for sustainable AI applications using drones'', co-supported by Exprivia S.p.A.~(CUP H91I22000410007).

\section*{Declaration on Generative AI}

During the preparation of this work, the authors used ChatGPT, Gemini, and Grammarly to assist with text refinement, organization, and language editing. After using these tools, the authors reviewed, verified, and edited all generated content, and take full responsibility for the accuracy and integrity of the final manuscript.

\bibliographystyle{elsarticle-num} 
\bibliography{main}
\end{document}